# Optimizing Software Effort Estimation Models Using Firefly Algorithm

**Nazeeh Ghatasheh[1], Hossam Faris[2], Ibrahim Aljarah[2], Rizik M. H. Al-Sayyed[2]**

[1]Department of Business Information Technology, The University of Jordan, Aqaba, Jordan
[2]Department of Business Information Technology, The University of Jordan, Amman, Jordan
Email: n.ghatasheh@ju.edu.jo, hossam.faris@ju.edu.jo, i.aljarah@ju.edu.jo, r.alsayyed@ju.edu.jo

## Abstract

**Software development effort estimation is considered a fundamental task for software development life cycle as well as for managing project cost, time and quality. Therefore, accurate estimation is a substantial factor in projects success and reducing the risks. In recent years, software effort estimation has received a considerable amount of attention from researchers and became a challenge for software industry. In the last two decades, many researchers and practitioners proposed statistical and machine learning-based models for software effort estimation. In this work, Firefly Algorithm is proposed as a metaheuristic optimization method for optimizing the parameters of three COCOMO-based models. These models include the basic COCOMO model and other two models proposed in the literature as extensions of the basic COCOMO model. The developed estimation models are evaluated using different evaluation metrics. Experimental results show high accuracy and significant error minimization of Firefly Algorithm over other metaheuristic optimization algorithms including Genetic Algorithms and Particle Swarm Optimization.**

## Keywords

**Software Quality, Effort Estimation, Metaheuristic Optimization, Firefly Algorithm**

## 1. Introduction

Effort estimation of software development has been a crucial task for software engineering community. Reliable effort estimation makes it more dependable to schedule project activities, allocate resources, estimate costs, and reduce the probability of project failures or delays. According to the survey in [1], most of the projects face overruns of effort or schedules. The survey also claimed that the lack of accurate estimation models is a main reason for project overruns.

Usually projects seem to be vague at the beginning and become less vague as they progress. At the same time, each project has its special nature that makes it much harder to estimate the required effort for completion. Due to the uncertain nature of projects, authors in [2] [3] suggested developing models that can adapt to a wide range of projects. But for the fact that software project data sets are typically small and the underlying relations are inaccurate or missing, the task of prediction becomes more challenging.

Several effort estimation models have been developed and improved over time for better prediction accuracy and thus better development quality [1] [4]-[8]. Such models range from complex calculations and statistical analysis of project parameters, to advanced machine learning approaches.

Heuristic optimization [9] is a method that relies on several attempts to find an optimal solution. Heuristic optimizers have been used in software effort estimation [10] as the use of genetic programming in [11] for model optimization. Another example is the part that Particle Swarm Optimization took in [12] as a heuristic optimizer. Moreover, the hybrid approaches encompass a combination of heuristic algorithms like the use of Genetic Algorithm and Ant Colony [13].

Despite a large number of experiments on finding the best prediction model, there is no clear evidence of a highly accurate or efficient approach. At the same time it is important to develop a prediction method that is less complex and much more useful. For instance, in some prediction models, a large number of variables that are used to construct the model do not reflect or improve the accuracy of the prediction model. Thus, collecting extra or unrelated variables is time-consuming with no significance. It would be more efficient to build a model with a minimum number of variables, hopefully finding the most important and common variables for generic project development efforts.

This work presents a study of how Firefly Algorithm improves the overall estimation of the software effort estimation. Where the main contributions are:
- Proving the suitability of Firefly Algorithm as predictor towards a generic prediction model for software effort estimation.
- The significant improvement in performance over previously reported methods.
- The suitability of machine learning approaches for effort prediction using a small number of input variables and data set instances.

## 2. Related Work

Many of Machine Learning (ML) approaches in the literature have been applied to improve the software effort estimation [2]. ML optimization algorithms that are inspired from nature have received much attention to find more accurate estimation for software effort. Nature-inspired ML algorithms include Cuckoo Search [14], Particle Swarm Optimization (PSO) [15], Bat Algorithm [16], Firefly Algorithm [17], and many others.

In [18], the authors compared the performance of different soft computing techniques such as PSO-Tuned COCOMO, Fuzzy Logic with traditional effort estimation structures. Their results showed that the proposed model outperformed traditional effort estimation structures for NASAs software effort data set. In [7], decision trees based algorithm was used to perform the software effort estimation. In addition, the authors presented an empirical proof of performance variations for several approaches that include Linear Regression, Artificial Neural Networks (ANN), and Support Vector Machines (SVM). Also the authors pointed to the suitability of the experimented ML approaches in the area of effort estimation. From their performance comparison results with other traditional algorithms, their results in terms of the error rate were better than other techniques.

A hybrid approach was adopted in [19] for parameter selection and model optimization. The authors used Genetic Algorithms (GA) for optimizing a Support Vector Regression model. The authors clarified the impact of using GA in feature selection and parameter optimization of the effort estimation model. The results of their approach showed that GA is applicable to improve the performance of the SVR model compared to other approaches. A generic framework is proposed in [20] for software effort estimation. The framework tries to simulate the human way of thinking to resolve the effort estimation by adopting fuzzy rules modeling. Therefore the generated models take advantage of experts knowledge, interoperable, and could be applied to various problems as risk analysis or software quality prediction. ANNs gained noticeable attention by researchers for effort estimation as illustrated by the review in [21], but it is insufficient to generalize the applicability of ANN in effort estimation. The authors stated that it is required to have further thorough investigation. The authors in [22] relied on seven evaluation measures to assess the stability of 90 software effort predictors over 20 data sets. According to the empirical results it was found that analogy-based methods or regression trees outperformed in terms of stability. Such conclusions open the door for extensive research towards a superior and generic prediction approach regarding the software effort estimation issue.

## 3. Firefly Algorithm

Firefly Algorithm (FA) is a multimodal optimization algorithm, which belongs to the nature-inspired field, is inspired from the behavior of fireflies or lightning bugs [17]. FA was first introduced by Xin-She at Cambridge University in 2007 [17]. FA is empirically proven to tackle problems more naturally and has the potential to over-perform other metaheuristic algorithms.

FA relies on three basic rules, the first implies that all fireflies are attracted to each other with disregard to gender. The second rule states that attractiveness is correlated with brightness or light emission such that bright flies attract less bright ones, and for absence of brighter flies the movement becomes random. The last main rule implies that the landscape of the objective function determines or affects the light emission of the fly, such that brightness is proportional to the objective function.

**Algorithm 1.** Pseudo-code of firefly algorithm.

```
Objective function  f(x)
x = (x1,···,xd)T
Generate initial population of fireflies  x_i (i = 1,2,···,n)
Light intensity  I_i  at  x_i  is determined by  f(x_i)
Define light absorption coefficient  γ
while (t < MaxGeneration) do
  for  i = 1: n  all  n  fireflies do
    for  j = 1: i  all n fireflies do
      if ( I_j > I_i ) then
        Move firefly i towards j in d-dimension;
      end if
      Attractiveness varies with distance  r  via  exp[γr]
      Evaluate new solutions and update light intensity
    end for
  end for
  Rank the fireflies and find the current best
end while
Post-process results and visualization
```

The attractiveness among the flies in FA has two main issues that are; the modeling of attractiveness and the various light intensities. For a specific firefly at location $X$ brightness $I$ is formulated as $I(X) \, \alpha f(X)$. While attractiveness $\beta$ is proportional to the flies and is related to the distance $R_{i,j}$ between fireflies $i$ and $j$. Equation (1) shows the inverse square of light intensity $I(r)$ in which $I_0$ represents the light intensity at the source.

$$I(r) = I_0 e^{-\gamma r^2} \tag{1}$$

Assuming an absorption coefficient of the environment $\gamma$, intensity is represented in Equation (2) in which $I_0$ is the original intensity.

$$I(r) = \frac{I_0}{1 + \gamma r^2} \tag{2}$$

Generally the Euclidean distance is illustrated in Equation (3), which represents the distance between a firefly at location $X_i$ and another at location $X_j$. In which $X_{i,k}$ is the $k^{th}$ component of the spatial coordinate $X_i$.

$$R_{ij} = \|x_i - x_j\| = \sqrt{\sum_{k=1}^{d}(x_{i,k} - x_{j,k})^2} \tag{3}$$

A firefly *i* attracted to a brighter one *j* as illustrated in Equation (4) where attraction is represented by $\beta_{e^{\gamma r_{ij}^2}}(x_j - x_i)$, and $\alpha\left(rand - \frac{1}{2}\right)$ represents the randomness according to the randomization parameter $\alpha$.

$$x_i = x_i + \beta_{e^{\gamma r_{ij}^2}}(x_j - x_i) + \alpha\left(rand - \frac{1}{2}\right) \qquad (4)$$

Furthermore, variations of attractiveness are determined by $\gamma$ which on its turn affects the behavior and convergence speed of FA.

## 4. Effort Estimation Models

One of the Famous and widely used effort estimation models is the Constructive Cost Models COCOMO and its extension COCOMOII. COCOMO is used as cost, effort, and schedule estimation model in the process of planning new software development activity, also known as COCOMO 81. COCOMO was defined between the late 1970s and early 1980s [23]. Where COCOMOII is a later extension of the previously defined model. This research work tries to optimize the parameters of three variations of the COCOMO model. The first is the basic COCOMO model which is represented in Equation (5).

$$E = a_i (\text{KLOC})^{b_i} \qquad (5)$$

*E* is the effort in person-months, KLOC represents the thousand (K) lines of code included in a software project. Typically, the coefficient $a_i$ and the exponent $b_i$ are chosen based on COCOMO pre-set parameters that depend on the software project details.

The other two models are extensions of the basic COCOMO model which are proposed by A. Sheta in [24]. Both models consider the effect of methodologies (ME) as supposed to be linearly related to the software effort. These models are represented in Equations (6) and (7) and named Model I and Model II respectively.

$$E = a_i (\text{KLOC})^{b_i} + c_i (\text{ME}) \qquad (6)$$

$$E = a_i (\text{KLOC})^{b_i} + c_i (\text{ME}) + d_i \qquad (7)$$

This work tries empirically to optimize the constants $a_i$, $b_i$, $c_i$ and $d_i$ using FA, GA and PSO.

## 5. Data Set and Evaluation Measures

This research considers a famous and public data set in order to produce comparable results; namely NASA projects' effort data set. The data set is challenging due to the small number of instances and limited number of analyzed variables. However, regarding the objectives of this research the data set is considered to be adequate. The data set is split into two parts; training set of about 60% and testing set of about 30% instances.

NASA data set [6] consists of 18 software projects for which this research considers three main variables that are the project size in thousand Lines of Code (KLOC), Methodology (ME), and Actual Effort (AE). Training data set has 13 instances and the records from 14 till 18 are for testing the model. **Table 1** shows the actual values of the training and testing data sets.

In order to check the performance of the developed models, the computed measures are the Correlation Coefficient ($R^2$),

$$R^2 = \frac{\sum_{i=1}^{n}(y_i - \bar{Y}_i)^2 - \sum_{i=1}^{n}(y_i - \hat{y}_i)^2}{\sum_{i=1}^{n}(y_i - \bar{y}_i)^2} \qquad (8)$$

the Mean Squares Error (MSE),

$$\text{MSE} = \frac{1}{n}\sum_{i=1}^{n}(y - \hat{y})^2 \qquad (9)$$

Table 1. NASA data set.

| Project No. | KDLOC | ME | Measured Effort |
|---|---|---|---|
| 1 | 90.2 | 30 | 115.8 |
| 2 | 46.2 | 20 | 96 |
| 3 | 46.5 | 19 | 79 |
| 4 | 54.5 | 20 | 90.8 |
| 5 | 31.1 | 35 | 39.6 |
| 6 | 67.5 | 29 | 98.4 |
| 7 | 12.8 | 26 | 18.9 |
| 8 | 10.5 | 34 | 10.3 |
| 9 | 21.5 | 31 | 28.5 |
| 10 | 3.1 | 26 | 7 |
| 11 | 4.2 | 19 | 9 |
| 12 | 7.8 | 31 | 7.3 |
| 13 | 2.1 | 28 | 5 |
| 14 | 5 | 29 | 8.4 |
| 15 | 78.6 | 35 | 98.7 |
| 16 | 9.7 | 27 | 15.6 |
| 17 | 12.5 | 27 | 23.9 |
| 18 | 100.8 | 34 | 138.3 |

the Mean Absolute Error (MAE),

$$\text{MAE} = \frac{1}{n}\sum_{i=1}^{n}|y_i - \hat{y}_i| \qquad (10)$$

the Mean Magnitude of Relative Error (MMRE),

$$\text{MMRE} = \frac{1}{n}\sum_{i=1}^{n}\frac{|y_i - \hat{y}_i|}{y_i} \qquad (11)$$

and the Variance-Accounted-For (VAF).

$$\text{VAF} = \left[1 - \frac{\text{var}(y(t) - \hat{y}(t))}{\text{var}(y(t))}\right] \times 100\% \qquad (12)$$

These performance criteria are used to measure how close the predicted effort to the actual values, where $y$ is the actual value, $\hat{y}$ is the estimated target value, and $n$ is the number of instances.

## 6. Experiments and Results

The experiments apply FA, GA and PSO for optimizing the coefficients of the basic COCOMO model, COCOMO Model I and COCOMO Model II based on the training part of NASA data set. For FA, the Matlab implementation developed by X.-S. Yang [9] is applied. Number of flies, particles and population size is unified and set to 100 in all the algorithms. The number of iterations is set to 500. The rest of the parameters of FA, GA and PSO are set as listed in **Tables 2-4**. MAE criteria are used as an objective function which is shown in Equation (10). In order to carry out meaningful evaluation results, each algorithm is applied 25 times then the average of the evaluation results is reported. In each run, the optimized models are evaluated based on the testing data using VAF, MSE, MAE, MMRE, RMSE and $R^2$ evaluation metrics.

Carrying out the experiments, the average convergence curves for FA, GA and PSO are shown in **Figures 1-3** respectively for the three variations of COCOMO model.

**Table 2.** Firefly algorithm parameter settings.

| Parameter | Value |
|---|---|
| Maximum iterations | 500 |
| Number of fireflies | 100 |
| Alpha | 0.4 |
| Betamin | 1 |
| Gamma | 0.4 |

**Table 3.** GA parameter settings.

| Parameter | Value |
|---|---|
| Maximum iterations | 500 |
| Population size | 100 |
| Selection method | Tournament selection |
| Crossover probability | 80% |
| Mutation probability | 5% |

**Table 4.** PSO parameter settings.

| Parameter | Value |
|---|---|
| Maximum iterations | 500 |
| Particles | 100 |
| Acceleration constant | [2.1, 2.1] |
| Inertia weight | [0.9, 0.6] |
| Maximum velocity | 100 |

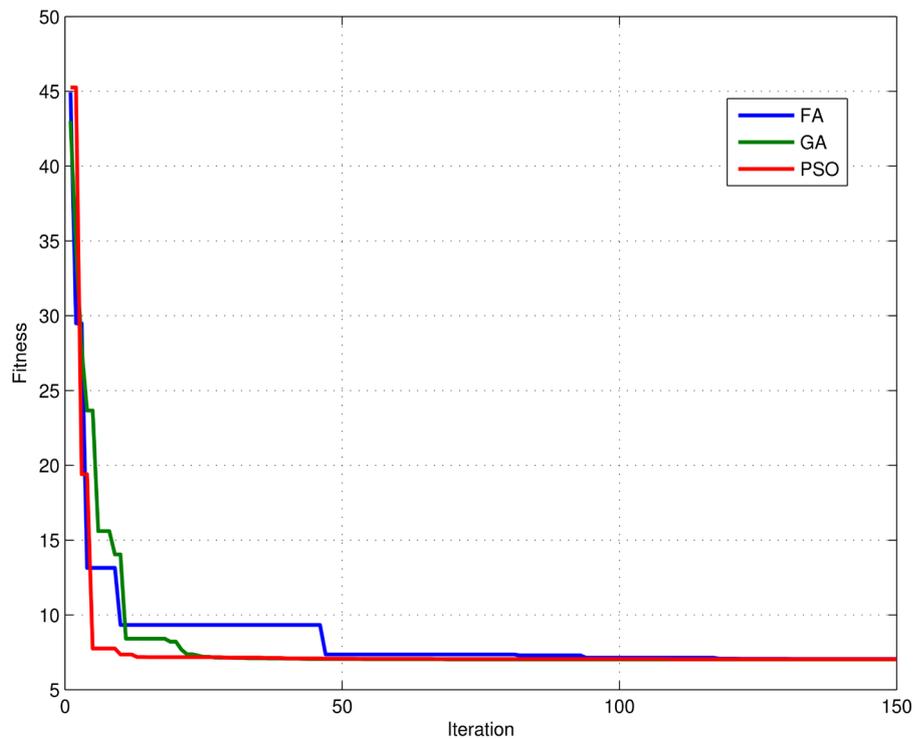

**Figure 1.** Convergence of FA, GA and PSO in optimizing the basic COCOMO model.

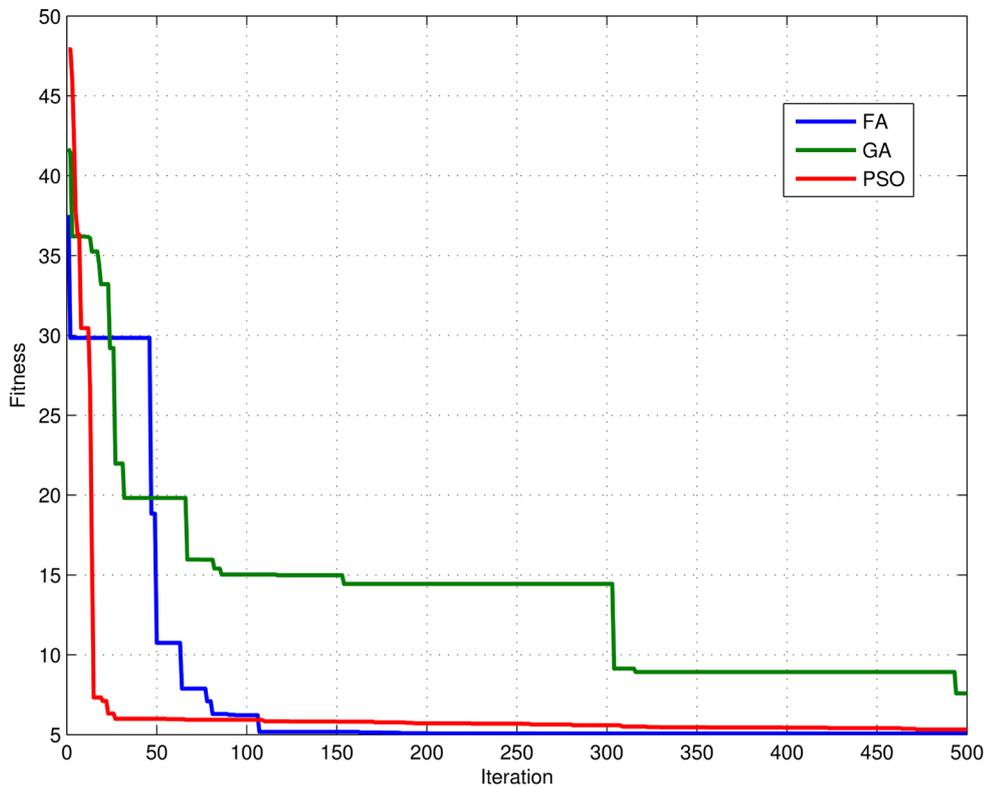

**Figure 2.** Convergence of FA, GA and PSO in optimizing Model I.

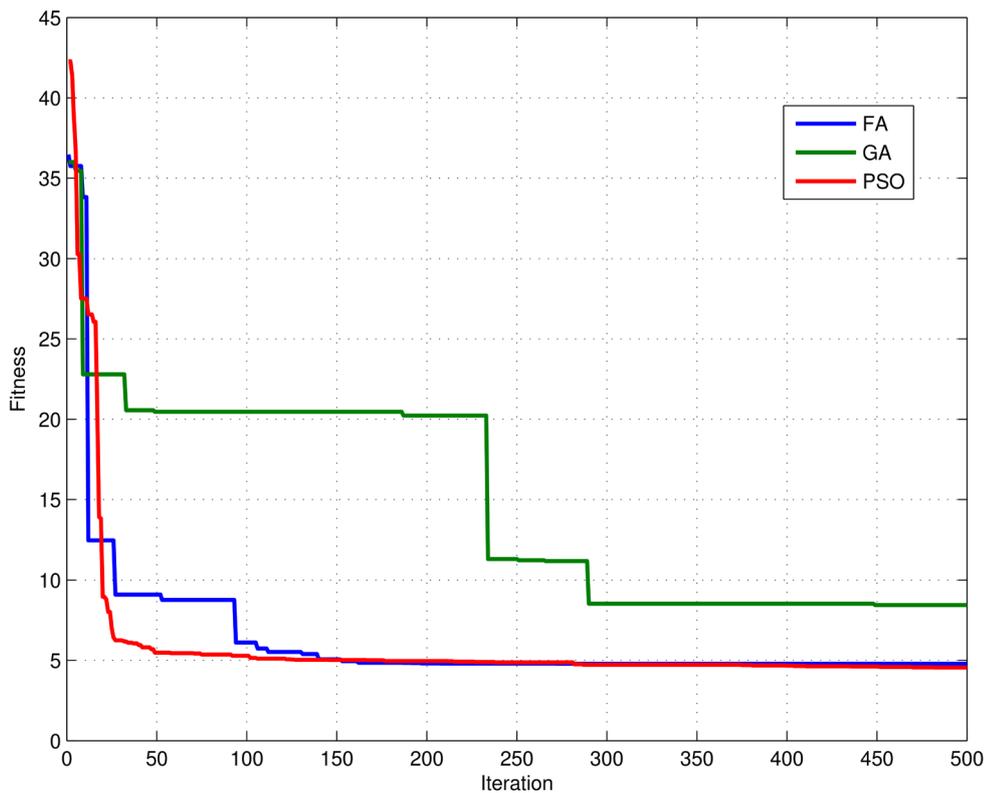

**Figure 3.** Convergence of FA, GA and PSO in optimizing Model II.

The evaluation results for training and testing cases are shown in **Tables 5-7**. Based on **Table 5** and **Table 6** it can be noticed that Firefly outperforms GA and PSO in optimizing the basic COCOMO model and the Model I by means of all evaluation metrics. For the Model II, Firefly and PSO are very competitive and have very close results. On the other hand GA has the lowest results and it has the slowest convergence.

In summary, FA as a metaheuristic optimization algorithm over-performs GA and PSO in terms of higher estimation accuracy for the software effort COCOMO based models.

**Table 5.** Basic COCOMO model.

|  | Training | | | Testing | | |
| --- | --- | --- | --- | --- | --- | --- |
|  | Firefly | GA | PSO | Firefly | GA | PSO |
| VAF | 93.82% | 93.72% | 93.73% | 98.16% | 97.97% | 97.98% |
| MSE | 104.88 | 107.28 | 107.15 | 59.14 | 63.96 | 63.68 |
| MAE | 7.04 | 7.03 | 7.03 | 5.65 | 6.06 | 6.04 |
| MMRE | 0.24 | 0.24 | 0.24 | 0.11 | 0.13 | 0.12 |
| RMSE | 10.24 | 10.36 | 10.35 | 7.67 | 8.00 | 7.98 |
| $R^2$ | 0.9367 | 0.9352 | 0.9353 | 0.9781 | 0.9763 | 0.9765 |

**Table 6.** COCOMO Model I.

|  | Training | | | Testing | | |
| --- | --- | --- | --- | --- | --- | --- |
|  | Firefly | GA | PSO | Firefly | GA | PSO |
| VAF | 96.78% | 92.94% | 96.96% | 98.62% | 97.97% | 98.52% |
| MSE | 56.05 | 127.70 | 54.16 | 47.74 | 98.17 | 60.07 |
| MAE | 5.42 | 8.94 | 5.16 | 5.56 | 7.70 | 5.63 |
| MMRE | 0.41 | 0.53 | 0.39 | 0.24 | 0.29 | 0.23 |
| RMSE | 7.48 | 10.95 | 7.36 | 6.82 | 9.39 | 7.72 |
| $R^2$ | 0.9662 | 0.9229 | 0.9673 | 0.9823 | 0.9637 | 0.9778 |

**Table 7.** COCOMO Model II.

|  | Training | | | Testing | | |
| --- | --- | --- | --- | --- | --- | --- |
|  | Firefly | GA | PSO | Firefly | GA | PSO |
| VAF | 96.95% | 92.42% | 97.48% | 98.63% | 97.60% | 98.70% |
| MSE | 53.74 | 129.37 | 45.28 | 45.02 | 114.79 | 52.85 |
| MAE | 5.36 | 8.20 | 4.43 | 5.57 | 7.83 | 5.29 |
| MMRE | 0.38 | 0.40 | 0.30 | 0.24 | 0.27 | 0.21 |
| RMSE | 7.26 | 11.05 | 6.72 | 6.62 | 9.86 | 7.19 |
| $R^2$ | 0.9676 | 0.9219 | 0.9727 | 0.9833 | 0.9575 | 0.9805 |

## 7. Conclusion and Future Work

This work investigated the efficiency of applying the Firefly Algorithm as a metaheuristic optimization technique to optimize the parameters of different effort estimation models. These models are three variations of the Constructive Cost Model COCOMO which are the basic COCOMO model, and other two extensions of the basic model that were proposed previously in the literature. The optimized models are assessed according to different evaluation criteria and compared with models optimized using other metaheuristic algorithms which are Genetic Algorithm and Particle Swarm Optimization. Evaluation results show that developed models using the Firefly Algorithm have higher accuracy in estimating software effort. Further future work is intended to overcome the instability issues, a more generic prediction model that is not highly affected by the size and the type of data set, and preferably an enhancement to the Firefly Algorithm itself. Moreover, it would be important to work towards a hybrid approach that encompasses the best characteristics of different prediction schemes.